\documentclass[conference]{IEEEtran}

\usepackage[pdftex]{graphicx}
\usepackage[ruled,vlined,linesnumbered]{algorithm2e}
\usepackage{array}
\usepackage{cite}
\usepackage{url}

\begin{document}

\newenvironment{conditions}
  {\par\vspace{\abovedisplayskip}\noindent\begin{tabular}{>{$}l<{$} @{${}={}$} l}}
  {\end{tabular}\par\vspace{\belowdisplayskip}}

\title{EDEN: Evolutionary Deep Networks for Efficient Machine Learning}

\author{\IEEEauthorblockN{Emmanuel Dufourq}
\IEEEauthorblockA{African Institute for Mathematical Sciences\\Maths \& Applied Maths, University of Cape Town \\
Email: edufourq@gmail.com}
\and
\IEEEauthorblockN{Bruce A. Bassett}
\IEEEauthorblockA{African Institute for Mathematical Sciences \\ South African Astronomical Observatory \\ Maths \& Applied Maths, University of Cape Town\\
Email: bruce.a.bassett@gmail.com}
}

\maketitle

\begin{abstract}
Deep neural networks continue to show improved performance with increasing depth, an encouraging trend that implies an explosion in the possible permutations of network architectures and hyperparameters for which there is little intuitive guidance. To address this increasing complexity, we propose Evolutionary DEep Networks (EDEN), a computationally efficient neuro-evolutionary algorithm which interfaces to any deep neural network platform, such as TensorFlow. We show that EDEN evolves simple yet successful architectures built from embedding, 1D and 2D convolutional, max pooling and fully connected layers along with their hyperparameters.   Evaluation of EDEN across seven image and sentiment classification datasets shows that it reliably finds good networks -- and in three cases achieves state-of-the-art results -- even on a single GPU, in just 6-24 hours. Our study provides a first attempt at applying neuro-evolution to the creation of 1D convolutional networks for sentiment analysis including the optimisation of the embedding layer.  
\end{abstract}


\IEEEpeerreviewmaketitle

\section{Introduction and Rationale}

Deep neural networks are powerful but unintuitive beasts whose wrangling requires experience,  significant trial and error to achieve good performance. The performance of such networks continued to improve as the depth is increased, e.g. \cite{Srivastava:2015:Training}. This along with the rising influence of deep learning in all fields means it is becoming more and more important to develop methods to automatically design optimal or near-optimal network architectures and hyperparameters. Deciding on the exact nature and order of the layers, choice of activation functions, number of units in fully connected layers, number of filters in convolutional layers and  other variables in creating deep neural networks is non-trivial. Given huge computing resources it is possible to simply try a vast number of possible combinations. Is there a way to be competitive with only a small amount of computing power, such as a single GPU? 

One solution, which we pursue here, is to evolve good neural networks through the use of evolutionary algorithms \cite{Sher:2012:Handbook}. Such neuro-evolutionary algorithms are not new, spanning nearly three decades, see e.g. \cite{Zhang:1995:Balancing}, \cite{Arifovic:2001:UsingGAs}, \cite{Idrissi:2016:GA}, beginning with a study that evolved the weights of the neural network\cite{Miller:1989:Designing}. 

Here we briefly summarise recent related work on neuro-evolutionary algorithms, which, by contrast to this study, have used very significant computing resources. Real et al. \cite{real:2017:large} proposed a neuro-evolutionary approach to optimise neural networks for image classification problems using a parallel system executed on 250 computers and achieved considerable success on the CIFAR image problems. Zoph and Le \cite{zoph:2016:neural} instead use recurrent neural networks along with reinforcement learning to learn good architectures. Eight hundred networks were trained on 800 GPUs.

Miikkulainen et al. propose CoDeepNEAT  \cite{miikkulainen:2017:evolving} in which a population of modules and blueprints are evolved. The blueprints are made up of several nodes which point to particular modules representing neural networks. Thus their proposed approach allows for the evolution of repetitive structures by enabling the blueprints to reuse evolved modules. 
Desell \cite{desell:2017:large} proposed EXACT, a neuro-evolutionary algorithm for deployment on a distributed cluster which they executed across 4500 volunteered computers and evolved 120,000 networks to tackle the MNIST dataset. Their approach did not use pooling layers and was limited to two dimensional input and filters. 

Finally we note that with a single GPU we have recently evolved deep networks to accurately identify whether a supervised machine learning challenge requires regression or classification \cite{Dufourq:2017:Automated}, achieving an average $96\%$ accuracy across a diverse set of tasks. This is a direct precursor of the current work and, given sufficient computing resources, can be seamlessly integrated into the network optimisation we discuss here. 

In this work we propose Evolutionary DEep Networks (EDEN), a neuro-evolutionary algorithm that combines the strengths of genetic algorithms  and deep neural networks to explore the search space of neural network architectures, their associated hyperparameters and the number of epochs to be applied. In our study, we explore additional features -- such as the optimisation of embedding layers -- and increase the complexity on the existing research. With EDEN we are interested in addressing two questions: can we evolve generally good architectures and hyperparameters for a broad range of problems (not just image classification)? Can this be successfully achieved on a single GPU, as opposed to the very large clusters used in previous studies?  

\begin{figure}[!h] 
  \centering
          \includegraphics[width=0.50\textwidth]{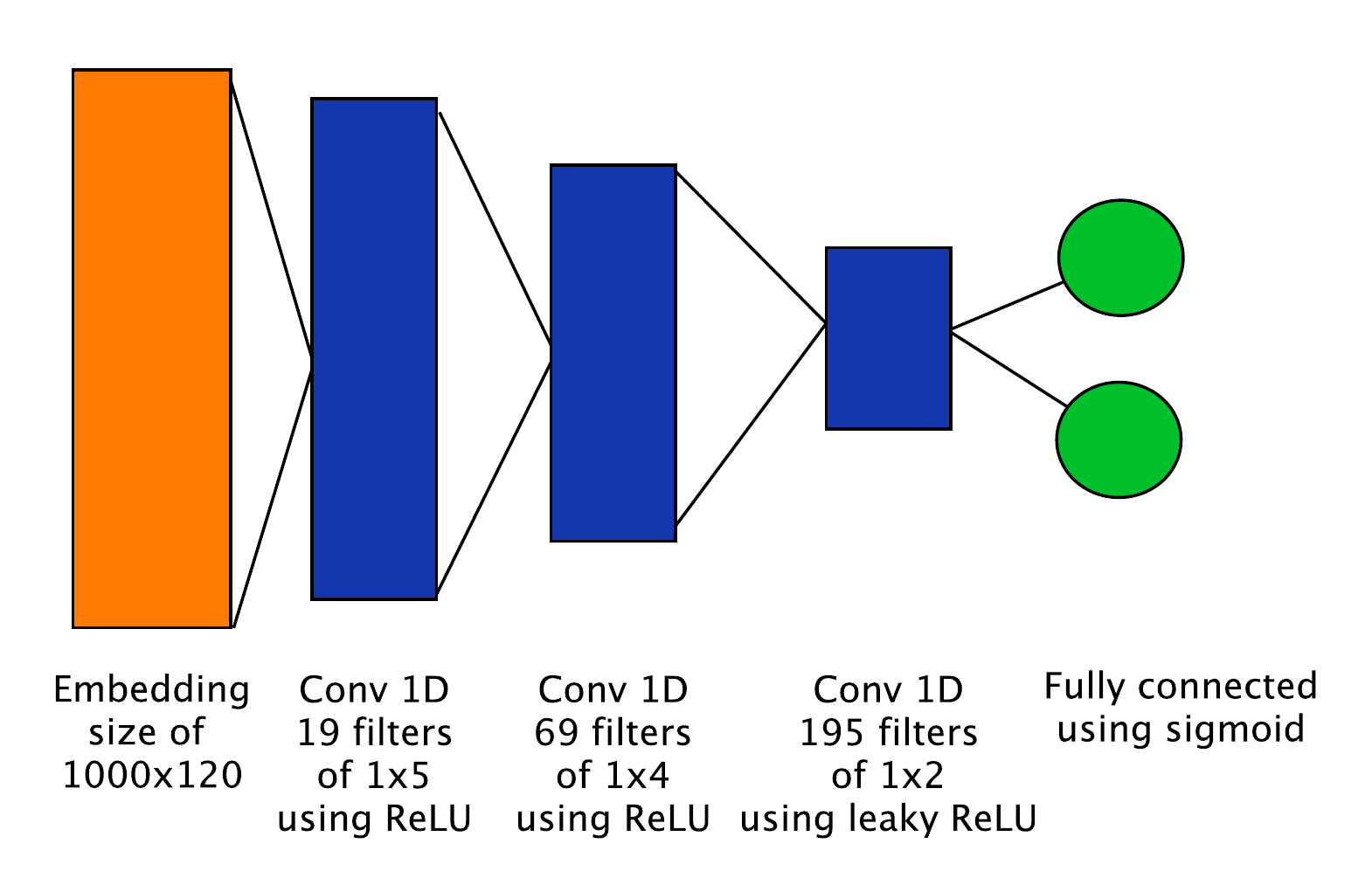}
  \caption{Each EDEN chromosome contains two genes, encoding the learning rate and a neural network. The figure illustrates an example of a neural network evolved using EDEN for a sentiment analysis task.  EDEN created an embedding layer with an output dimension of 120, followed by three 1D convolutional layers. EDEN evolved the number of filters, each filters' dimension along with each corresponding activation function. For the last layer, the selected activation function which EDEN determined was the sigmoid function. The learning rate for this chromosome is 0.0023.} \label{fig:eden}
\end{figure}

We interface EDEN to TensorFlow \cite{tensorflow:2015:whitepaper} and thus new layers, functions and other features can easily be incorporated and controlled by EDEN as these represent function calls to the respective TensorFlow functions. Additionally, EDEN is not limited to TensorFlow, other modern deep neural network platforms can be interfaced. Figure \ref{fig:eden} illustrates an example of a neural network architecture encoded by an EDEN chromosome.

The associated video\footnote{\url{https://vimeo.com/234510097}} illustrates the evolution of the chromosomes during the execution of EDEN on the MNIST image classification problem, showing the population converging towards an efficient solution made up primarily of two-dimensional convolutional layers.

\section{Genetic Algorithm}\label{sec: gas}

A genetic algorithm (GA) \cite{Goldberg:1989:GeneticAlgorithms} is an evolutionary algorithm which can be applied to solve optimisation problems. A population of chromosomes is randomly initialised. Each chromosome represents a candidate solution to the optimisation problem. A fitness function is used to evaluate each chromosome to determine the extent to which the chromosome can solve the problem. In a generational model, the GA iterates multiple times, known as generations, until some predetermined condition is met (for example, a maximum number of generations). Each chromosome is made up of several genes, and these genes are altered using a genetic operator. The resulting chromosome after the application of a genetic operator is known as an offspring. Multiple offspring are created based on the population size. The offspring replace the current chromosome population in each generation. In this study we used the traditional GA. We additionally increment the number of neural network epochs along with the number of generations to explore the best value for the number of epochs. Algorithm \ref{algo:ga-algo} presents the GA used.

\begin{algorithm}

\SetKwData{Gen}{generation}
\SetKwData{Epoch}{epochs}
\SetKwData{PopS}{population\_size}
\SetKwData{Genmax}{generation\_max}
\SetKwInOut{Input}{input}

\Input{\Epoch : number of neural network epochs}
\Input{\PopS : population size}
\Input{\Genmax : maximum number of GA generations}

	\Begin{

	\Gen $\leftarrow 0$.

	\Epoch $\leftarrow$ \Epoch.

	\PopS $\leftarrow$ \PopS.

	Create an initial population of chromosomes.
	
	Evaluate the initial population.
	
	\While{\Gen $\leq$ \Genmax}{
	
		\If{\Gen $\neq$ 0}{

			\Epoch $\leftarrow$ (\Epoch $+ 1$).

			\PopS $\leftarrow$ (\PopS $- 10$).

		}
	
		Select the parents.
		
		Create offspring using the genetic operators.
		
		Replace the current population with the new offspring created in step 12.

		Evaluate the current population.

		\Gen $\leftarrow \Gen+1.$
	
	}

\Return{The best chromosome.}

}
\caption{Modified genetic algorithm used in this study} 
\label{algo:ga-algo} 
\end{algorithm}

We choose to use GAs since the complexity of the chromosomes can be increased or decreased based on the number of genes which are encoded. GAs provide a further key advantage over other optimisation algorithms: they fluently handle complex combinations of discrete (e.g. layer type) and continuous (e.g. learning rate) search spaces, making them ideal for neuro-evolutionary studies; e.g. \cite{real:2017:large, Dufourq:2017:Automated}.

\section{Proposed Chromosome}\label{sec:proposed}

Each EDEN chromosome is made up of two genes, and these genes constitute the required components to optimise a single neural network on some given input classification dataset. The two genes encode the learning rate and the network architecture. The learning rate denotes the value which is applied during the training optimisation. The architecture represents the exact order of the neural network layers and operations. 

\subsection{Network Layers}

The following layers and operations were made available to EDEN: two-dimension convolution \cite{lecun:1989:generalization}, one-dimension convolution \cite{kim:2014:convolutional}, fully connected, dropout \cite{Srivastava:2014:Dropout}, one-, and two-dimension max pooling \cite{Zhou:1988:Computation} and embedding \cite{Maas:2011:LearningWord}. Inappropriate choices (such as using a 2-D convolution for a text sentiment problem) are penalised as described in \cite{Dufourq:2017:Automated}.

For the sentiment analysis tasks, instead of using pre-trained vectors such as Word2Vec \cite{mikolov:2013:efficient}, or setting a pre-determined embedding dimension size, we decided to allow EDEN to learn the dimension of the word embeddings as part of the optimisation. We created a dictionary by mapping each unique words their frequency count in the training data. We took the top 1000 most frequent words and used this to encode the text into vectors of integers. Enabling EDEN to optimise both the vocabulary size and the embedding would result in significant computation time and hence this was not included in this study.

\subsection{Activation Functions}

When a layer is randomly generated an activation function is also randomly selected. Convolutional layers can choose between the following functions: \{linear, leaky relu, prelu, relu\}. Fully connected layers choose from: \{linear, sigmoid, softmax, relu\}. The last fully connected layer in the network can use any of: \{linear, sigmoid, softmax\}. These functions were selected as they are commonly used in literature. It is however possible to include a larger number of activations functions.

\section{EDEN} \label{sec:wizard}

\subsection{Initial Population Generation}

The first phase in an evolutionary algorithm is to create the initial population of chromosomes. These chromosomes denote the first generation of solutions to the optimisation problem, i.e. in this case to generate neural networks that can correctly classify data. The number of chromosomes to create in the initial population (initial population size) is a user-defined parameter. Once each chromosome is created it is evaluated to determine how close it is to the optimal solution (100\% classification accuracy), see section \ref{sec:fitness}. 

The initial population generation method used in this study was inspired by the ramped-half-and-half method proposed by Koza \cite{Poli:2008:AField} which enables the creation of candidate solutions of various sizes. In a similar manner, we implemented an initial population generation method that would create neural network architectures of various sizes to increase the amount of diversity in the initial population (as opposed to a population that is skewed towards a particular size). 

Algorithm \ref{algo:create-chromosome} outlines the pseudocode on how a chromosome was randomly generated and algorithm \ref{algo:create-population} presents the initial population generation method used in this study. In certain cases, invalid architectures can be generated, these invalid architectures are discarded and a new one is generated.

Given our computational limitations, we had to limit the search space by setting bounds on certain variables. Real et al. \cite{real:2017:large} did not implement these limitations, however it is worth noting that in their study they used 250 machines. The \textit{keep} probability for dropout was randomly generated between 0 and 1 as these are the only acceptable values.

The bounds for each variable are listed below.

\begin{itemize}
\item number of filters in 1D and 2D convolution: [10, 100]
\item filter size for 1D and 2D convolution: [1, 6]
\item kernel size for 1D and 2D max pooling: [1, 6]
\item number of units in fully connected layers: [10, 100]
\item embedding layer output size: [100, 300]
\end{itemize}

\begin{algorithm}[!h]

\SetKwData{PopS}{population\_size}
\SetKwInOut{Input}{input}

\Input{\PopS : population size}

\SetKwInOut{Input}{input}

	\Begin{
	
	\For{$i \gets 0$ \KwTo $\PopS$}{

		Generate chromosome with size = $(\lfloor\frac{i}{10}\rfloor + 1)$

		Determine number of parameters

		Evaluate chromosome's validation accuracy

		Add chromosome to initial population
	
	}

}
\caption{Creating initial population of chromosomes of various architecture sizes} 
\label{algo:create-population} 
\end{algorithm}

\begin{algorithm}

\SetKwData{size}{chromosome\_size}
\SetKwData{type}{layer\_type}
\SetKwData{drop}{dropout\_allowed}
\SetKwData{newlayer}{new\_layer}

\SetKwInOut{Input}{input}

\SetKwProg{Fn}{Function}{}{}
\Input{\size : maximum number of genes in chromosome}

	\Begin{
	
	Initialise an empty chromosome.

	\type $\leftarrow$ `cnn'

	\For{$i \gets 0$ \KwTo $\size - 1$}{

		\If{$i = 0$}{
			\drop $\leftarrow$ false
		}
		\Else{
			\drop $\leftarrow$ true
		}

		\newlayer $\leftarrow$ $CreateLayer(\drop, \type)$

		Append $newlayer$ to chromosome

		\If{$newlayer$ is fully connected}{
			\type $\leftarrow$ `non-cnn'
		}

	}

	Randomly create fully connected layer and append to chromosome

\Return{chromosome.}

}

\Fn{CreateLayer ($dropout$, $type$)}{

	\If{$type = `cnn'$}{
		\If{$dropout = true$}{
			Randomly create convolution, fully connected or dropout operation 
		}
		\Else{
			Randomly create convolution layer
		}
	}
	\Else{
		\If{$dropout = true$}{
			Randomly create fully connected layer or dropout operation 
		}
		\Else{
			Randomly create fully connected layer
		}

	}
}
\caption{Creating an EDEN chromosome.} 
\label{algo:create-chromosome} 
\end{algorithm}

\subsection{Parent Selection}

During each generation of the GA, parents must be selected to create offspring using a genetic operator. Parents are obtained using a parent selection method. Three common parent selection methods are fitness-proportionate, rank and tournament selection \cite{Blickle:1996:AComparison}. For this study, tournament selection (algorithm \ref{algo:tournament-selection}) was used given that it was shown to be a successful method by Zhong et al. \cite{Zhong:2005:ComparisonOfPerformance}.

The algorithm works as follows. A number (tournament size) of chromosomes are randomly selected from the current population. The tournament size is a user-defined parameter. Once the chromosomes have been randomly selected they are each evaluated using the fitness function. The chromosome with the smallest fitness (a smaller fitness denotes a better performing chromosome since the validation error is used in the computation of the fitness) is then returned as the parent to be used by the genetic operator. 

\subsection{Mutation}

The recombination genetic operator was not included in our study,  similar to Real et al. \cite{real:2017:large}. For each execution of the mutation operator a single parent is obtained using tournament selection. The mutation operator is applied to the parent to generate offspring\textsubscript{1}. The mutation operator is then applied to offspring\textsubscript{1} and consequently creates offspring\textsubscript{2}. The fitness of offspring\textsubscript{1}, offspring\textsubscript{2} and the original parent chromosome is compared. The chromosome with the lowest fitness is returned and placed into the new population. Preliminary experiments revealed that performing mutation once on a parent prohibited the algorithm from sufficiently exploring the search space. It was for this reason that we repeated the mutation operator to generate two variations of offspring.

The details about how the mutation operator changes a chromosome are as follows. For a given chromosome, the operator randomly changes either the chromosome's learning rate or the neural network layers. In the case that the learning rate is selected, then a new value for the learning rate is randomly generated as was discussed in section \ref{sec:proposed}. In the case where the neural network layers is selected, then the operator either adds a new layer, deletes a layer or replaces one. The choice is made based on the size of the architecture. If the size of the chromosome's architecture has reached the maximum size (predetermined), then a layer can either be deleted or replaced. However, if the size is less than the maximum allowed size then a layer can either be added, deleted or replaced. A constraint was put in place so that the mutation operator cannot remove the first or last layers. 

\textit{Deletion} is performed by randomly selecting any layer (excluding the first or last layers) and removing it from the network architecture in the chromosome. \textit{Replacement} is performed by randomly selecting a layer within the architecture and removing it. An entirely new layer is generated and inserted in the same position as the one which was removed. \textit{Addition} generates a new layer and adds it anywhere in the architecture.

It is possible that the randomness within the mutation results in invalid neural network architectures. After each application of the mutation operator, a check is performed to assess the validity of the resulting architecture. If mutation generates an invalid architecture, then the mutation operator is applied again until a valid one is generated.

The number of neural network parameters is computed for each offspring created. The parameters represent the number of trainable weights in the neural network. Larger values denote more complex models, and small numbers consequently denote less complex ones.

\begin{algorithm}[!ht]

\SetKwData{Size}{size}
\SetKwData{Current}{current\_best}
\SetKwData{Chromosome}{random\_chromosome}
\SetKwInOut{Input}{input}
\SetKwInOut{Output}{output}

\Input{\Size : size of the tournament}
\Output{The best chromosome which will be used as a parent}

\Begin{

	\Current $\leftarrow$ null
	
	\For{$i\leftarrow 1$ \KwTo \Size}{
		
		\Chromosome $\leftarrow$ randomly select a chromosome from the population

		Evaluate \Chromosome

		\If{fitness of \Chromosome $<$ fitness of \Current}{

			\Current $\leftarrow$ \Chromosome

		}

	}

\Return{\Current}

}
\caption{Pseudocode for tournament selection.} 
\label{algo:tournament-selection} 
\end{algorithm}

\subsection{Chromosome Evaluation} \label{sec:fitness}

A fitness function is required to steer EDEN towards an optimal solution. This function computes a fitness score -- a numerical value which denotes how `good' a chromosome is. For this study, we chose a fitness function that makes use of the error on the validation set (the dataset was split into training, validation and testing subsets) as well as the number of trainable parameters. The relative importance of these two is controlled by $\alpha$, a complexity parameter. This fitness function rewards less complex and more accurate models compared to more complex, less accurate ones. Furthermore, it helps to break ties when two chromosomes have the same validation error. We fix $\alpha = 1$, but this can be changed depending on the relative importance of performance versus the need for small networks in the problem at hand.  

\begin{equation}
{\mbox{fitness}(\mbox{Net}) = \mbox{val\textsubscript{error}} + \alpha \left(1- \frac{1}{\mbox{N\textsubscript{p}}}\right)} 
\end{equation}

where

\begin{conditions}
 \mbox{Net} & the neural network being evaluated \\
 \mbox{val\textsubscript{error}} & the validation error \\
 \mbox{N\textsubscript{p}}  & the number of trainable parameters \\ 
 \mbox{$\alpha$} & complexity parameter (default $\alpha = 1$)\\
\end{conditions}

Once the mutation operator generates an offspring, the architecture and hyperparameters which are encoded in the chromosome are used to train a neural network using TensorFlow. The training data is used in the training of the network. The categorical cross entropy loss function is used during training. Once the network is done training then the neural network is evaluated on the validation data using the fitness function. The fitness obtained from the function is then stored as the chromosome's fitness. The fitness function used in this study is presented in equation 1. 

\section{Experimental Setup} \label{sec:setup}

For each dataset, we executed EDEN 5 times and averaged the results (similar to \cite{real:2017:large}). EDEN was evaluated on a single machine with a MSI GeForce GTX1070 and 16GB of CPU RAM. During the evolutionary process an experiment used between 4GB to 7GB CPU RAM based on the dataset, and the GPU utilisation varied from 50 to 99 percent during the training of the neural networks. The algorithm was developed in Python 3.6.1, TensorFlow 1.2.1 and Keras 2.0.6 \cite{chollet:2015:keras}. Keras was used to determine the number of parameters for the neural networks contained in each chromosome. The operating system was Ubuntu 16.04 LTS. 

\subsection{Datasets} \label{sec:datasets}

Table \ref{table:dataset} presents the datasets for which EDEN was evaluated on. IMDB \cite{maas:2011:IMDB} and Electronics \cite{johnson:2014:effective} are sentiment analysis datasets. The other datasets -- namely, MNIST \cite{LeCun:1998:Gradient}, CIFAR-10 \cite{krizhevsky:2009:learning}, Fashion-MNIST \cite{Xiao:2017:Fashion-MNIST} and the two EMNIST datasets \cite{cohen:2017:emnist} -- were image classification problems. For each dataset, EDEN was trained on the training data, the validation set was used to evaluate the performance of the chromosomes and the test set was used when reporting the results. The training and testing split is presented in the table. The datasets did not contain any missing values, and the class values were converted into their respective one-hot encoded values.

\begin{table}[ht]
\caption{The 7 datasets used in this study. The number of training and testing samples are provided along with the number of classes for each dataset. The IMDB and Elec datasets are sentiment analysis problems, and the remaining datasets are image classification problems.} \label{table:dataset}
\begin{centering}
\begin{tabular}{|l|c|c|c|}
\hline 
\textbf{Dataset} & \textbf{Training} & \textbf{Testing} & \textbf{Classes}\tabularnewline
\hline 
CIFAR-10 & 50,000 & 10,000 & 10 \tabularnewline
\hline 
Elec & 25000 & 25000 & 2 \tabularnewline
\hline 
EMNIST - Balanced & 112,800 & 18,800 & 47 \tabularnewline
\hline 
EMNIST - Digits  & 240,000 & 40,000 & 10 \tabularnewline
\hline 
Fashion-MNIST & 60,000 & 10,000 & 10 \tabularnewline
\hline 
IMDB & 25000 & 25000 & 2 \tabularnewline
\hline 
MNIST & 60,000 & 10,000 & 10 \tabularnewline
\hline 

\end{tabular}
\par\end{centering}

\end{table}

\subsection{Parameters} \label{sec:params}

Table \ref{table:params} present the GA and neural network parameters used throughout this study. Preliminary runs were performed to obtain these values. The purpose of EDEN was to, amongst other things, evolve the neural network's hyperparameters and thus the parameters presented in the table were the only values which were input into EDEN.

\begin{table} [ht]
\caption{The GA and neural network parameters used in this study. We conducted additional experiments
to select these parameters. The number of generations was not set
to a high value to avoid extreme runtimes. The neural network parameters 
were used by EDEN during the training of the neural networks.} \label{table:params}
\begin{centering}
\begin{tabular}{|>{\centering}p{3.6cm}|>{\centering}p{3.7cm}|}
\hline 
\textbf{Parameter} & \textbf{Value}\tabularnewline
\hline 
Number of generations & 10 \tabularnewline
\hline 
Initial population size & 100 \tabularnewline
\hline 
Tournament size & 7 \tabularnewline
\hline 
Number of epochs & starting value of 3, incremented by 1 every generation\tabularnewline
\hline 
Weight initialisation - mean \& standard deviation & 0.00, 0.01\tabularnewline
\hline 
Batch size & 1024\tabularnewline
\hline 
Optimiser& Adam \cite{kingma2014adam}  \tabularnewline
\hline 
\end{tabular}
\par\end{centering}

\end{table}

\section{Results and Conclusion} \label{sec:conclusion}

Table \ref{table:results} presents the average test accuracy and number of trainable parameters. The best results (for the population size which we used) were obtained when using Adam. EDEN was initially configured to include the optimiser function in the chromosome, but the results revealed that this did not improve the results. It is possible that a larger population size would have yielded interesting results by searching for the most optimal network optimiser. 

\begin{table}
\caption{The average best EDEN test accuracy (\%) after 10 generations and 13 epochs of training (standard deviation shown in parentheses). The average number of trainable EDEN parameters and the previous state-of-the-art results and reference are also shown alone with the average learning rates (denoted LR) which were evolved for the best chromosomes.} \label{table:results}
\begin{centering}
\begin{tabular}{|l|>{\centering}p{1.15cm}|c|l|>{\centering}p{1.4cm}|}
\hline 
\textbf{Dataset} & \textbf{Test Accuracy} & \textbf{LR} & \textbf{Params} & \textbf{State of the art (\%)}\tabularnewline
\hline 
CIFAR-10 & 74.5 (3.1) & 0.0024 & 172,767 & \textbf{97.14} \cite{gastaldi:2017:shake}  \tabularnewline
\hline 
Elec & 87.2 (0.5) & 0.0040 & 26,625 & \textbf{93.17} \cite{Vishwanath:2016:AddingCNNs} \tabularnewline
\hline 
EMNIST-Bal. & \textbf{88.3} (0.8) & 0.0019 & 1,688,43 & 78.02 \cite{cohen:2017:emnist} \tabularnewline
\hline 
EMNIST-Digits  & \textbf{99.3} (0.1) & 0.0027 & 3,001,576 & 95.90 \cite{cohen:2017:emnist} \tabularnewline
\hline 
Fashion-MNIST & \textbf{90.6} (0.5) & 0.0059 & 4,624,447 & 89.7 \cite{Xiao:2017:Fashion-MNIST}  \tabularnewline
\hline 
IMDB & 85.8 (0.6) & 0.0053 & 319,185 & \textbf{93.34} \cite{Vishwanath:2016:AddingCNNs} \tabularnewline
\hline 
MNIST & 98.4 (0.3) & 0.0031 & 1,857,601 & \textbf{99.79} \cite{Wan:2013:Regularization} \tabularnewline
\hline 

\end{tabular}
\par\end{centering}

\end{table}
Table \ref{table:results} shows our results. 
EDEN achieved new state-of-the-art results on the EMNIST-balanced, EMNIST-digits and Fashion-MNIST datasets. For the two sentiment analysis tasks (Elec and IMDB) EDEN evolved neural networks which produced good -- but sub-state-of-the-art -- accuracy despite EDEN's ability to optimise the embedding layer. In future work, we will determine the effect of also allowing EDEN to optimise the vocabulary size. The average evolved learning rate ranged between 0.00186 and 0.0059. Average execution times, in hours, for a single EDEN experiment of ten generations were 9, 7, 18, 24, 12, and 6 for IMDB, Elec, EMNIST-balanced, EMNIST-digits, CIFAR-10 and Fashion-MNIST respectively.  In addition, we enforce the constraint that no networks receive more than 13 epochs of training. As a result EDEN took, on average, 12 hours for the MNIST dataset (accuracy 98.4\%); significantly less than the 2 month execution time of EXACT which achieved a similar accuracy of 98.32\% \cite{desell:2017:large}.

\begin{figure}[!h] 
  \centering
          \includegraphics[width=0.49\textwidth]{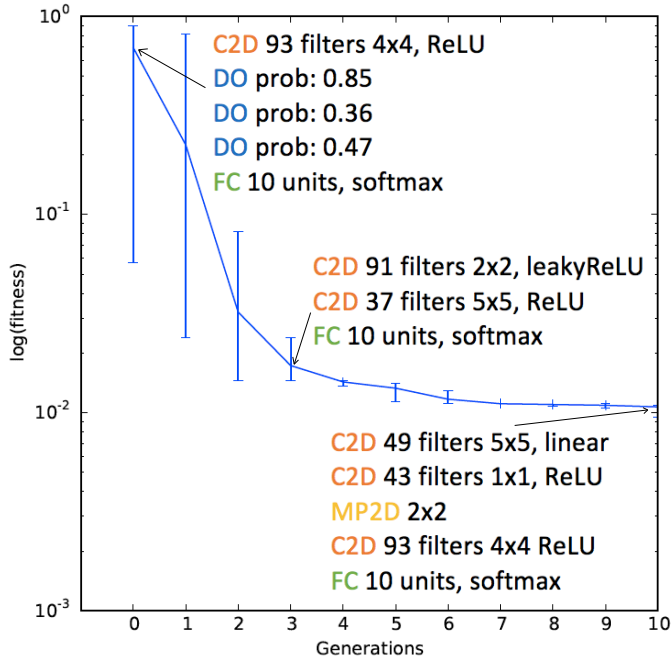}
  \caption{Illustrating the change in mean fitness over the GA generations for the MNIST data. Error bars show the 5\% and 95\% percentile values in fitness across the population. Initially there is significant variance in the fitness which reduces as the solutions improve and the population converges. We also show three networks sampled from the initial, mid-point and final generations, along with their associated hyperparameters. We show the best evolved network at three stages during the evolution. Here C2D, MP2D, DO, FC represent 2D convolution, 2D Max Pooling, Drop Out and Fully Connected layers respectively.} \label{fig:changefitness}
\end{figure}

\begin{figure}[!h] 
  \centering
          \includegraphics[width=0.50\textwidth]{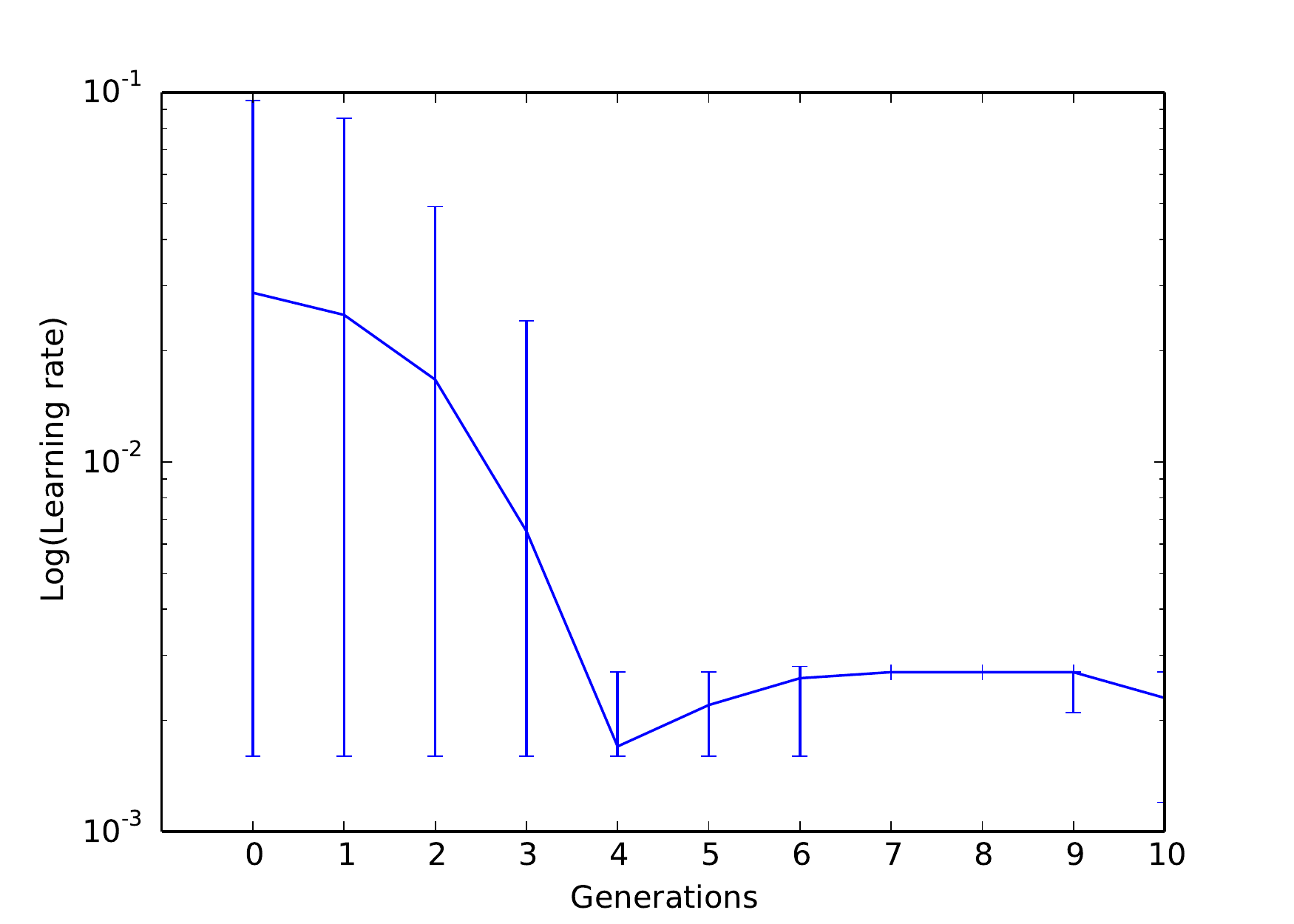}
  \caption{Change in mean learning rate over the GA generations. Error bars show the 5\% and 95\% percentile value in terms of the learning rate variance in the population. Initially the chromosomes are random so there is a lot of variance in the learning rate. This changes as the population converges towards better solutions.} \label{fig:changelearnrate}
\end{figure}

EDEN did not, however, produce competitive results on CIFAR-10, obtaining an average test accuracy of 74.5\% after 13 (80.5\% after 100 epochs) training epochs of the final network evolved after 12 hours, compared to the current 97.14\% state-of-the-art \cite{gastaldi:2017:shake}. This is primarily due to the 7-layer depth constraint we imposed due to running EDEN on a single GPU. As a result the best model that EDEN evolved had only 172,767 parameters, only $0.7\%$ of the 26.2 million used in the state-of-the-art \cite{gastaldi:2017:shake}. Figures \ref{fig:changefitness} and \ref{fig:changelearnrate} illustrate the change in fitness and learning rate over the evolutionary process. Both figures show the convergence of the population. The fitness rapidly decreases from the random initial population to generation 3, after which the fitness decreases at a slower rate.

Determining optimal or efficient deep neural network architectures and hyperparameters is a challenging task. Researchers and practitioners who are new to the creation of deep neural networks can benefit from algorithms which automatically create architectures and determine hyperparameters. In our study, we propose EDEN, a neuro-evolutionary algorithm that interfaces with TensorFlow -- or any other deep neural network platform -- to automatically create architectures and optimise hyperparameters. Here EDEN was evaluated on classification problems, but can easily be applied to regression problems.

EDEN is designed to evolve efficient deep networks and for each dataset was executed on a single GPU running for 24 hours or less. The findings reveal that competitive results can be obtained using significantly less computational power than has been deployed in other neuro-evolutionary studies. Evaluated on image classification and sentiment analysis problems, EDEN achieves state-of-the-art results in three of seven datasets. Our study is also a first attempt at applying neuro-evolution to the creation of 1D convolutional networks for sentiment analysis, optimising an embedding layer for sentiment analysis. In future work, we intend on extending EDEN to evolve generative adversarial networks architectures, as well as exploring parallel implementations.

\section*{Acknowledgment}
The financial assistance of the National Research Foundation (NRF) towards this research is hereby acknowledged. Opinions expressed and conclusions arrived at, are those of the authors and are not necessarily to be attributed to the NRF.

\bibliographystyle{IEEEtran}

\end{document}